\documentclass[nonblindrev]{informs3} 

\usepackage{balance}
\usepackage{amsmath}
\usepackage[tight,footnotesize]{subfigure}
\usepackage{graphicx}
\usepackage{bm}
\usepackage{algorithm}
\usepackage[noend]{algorithmic}
\usepackage{multirow}
\usepackage{slashbox}
\usepackage{caption}
\usepackage{amsfonts}

\OneAndAHalfSpacedXII 

\usepackage{natbib}
 \bibpunct[, ]{(}{)}{,}{a}{}{,}%
 \def\bibfont{\small}%
\TheoremsNumberedThrough     

\EquationsNumberedThrough    

\MANUSCRIPTNO{2020} 

\begin{document}


\RUNAUTHOR{Tang et al.}

\RUNTITLE{Beyond Pointwise Submodularity:   Non-Monotone Adaptive Submodular Maximization in Linear Time}

\TITLE{Beyond Pointwise Submodularity:   Non-Monotone Adaptive Submodular Maximization in Linear Time}

\ARTICLEAUTHORS{%
\AUTHOR{Shaojie Tang}
\AFF{Naveen Jindal School of Management, The University of Texas at Dallas}
} 

\ABSTRACT{In this paper, we study the non-monotone adaptive submodular maximization problem subject to a cardinality constraint.  We first revisit the  adaptive random greedy algorithm  proposed in \citep{gotovos2015non}, where they show that this algorithm achieves a $1/e$ approximation ratio if the objective function is adaptive submodular and pointwise submodular. It is not clear whether the same guarantee holds under adaptive submodularity (without resorting to pointwise submodularity) or not. Our first contribution is to show that the adaptive random greedy algorithm achieves a $1/e$ approximation ratio under  adaptive submodularity.  One limitation of the adaptive random greedy algorithm is that it requires $O(n\times k)$ value oracle queries, where $n$ is the size of the ground set and $k$ is the cardinality constraint. Our second contribution is to develop the first linear-time algorithm for the non-monotone adaptive submodular maximization problem. Our algorithm achieves a  $1/e-\epsilon$ approximation ratio (this bound is improved to $1-1/e-\epsilon$ for monotone case), using only $O(n\epsilon^{-2}\log \epsilon^{-1})$ value oracle queries.    Notably, $O(n\epsilon^{-2}\log \epsilon^{-1})$ is independent of the cardinality constraint.  For the monotone case, we propose a faster algorithm that achieves a $1-1/e-\epsilon$ approximation ratio in expectation  with $O(n \log  \frac{1}{\epsilon})$ value oracle queries. We also generalize our study by considering a partition matroid constraint, and develop a linear-time algorithm for monotone and fully adaptive submodular functions. 
}


\maketitle

\section{Introduction}
Maximizing a submodular function subject to practical constraints has been extensively studied in the literature \citep{golovin2011adaptive,tang2020influence,yuan2017no,krause2007near}. For the non-adaptive setting  where all items must be selected at once, \citet{nemhauser1978analysis} show that the greedy algorithm achieves a $1-1/e$ approximation ratio when maximizing a monotone submodular function subject to a cardinality constraint. Their algorithms performs $O(n\times k)$ value oracle queries, where $n$ is the size of the ground set and $k$ is the cardinality constraint.  Much research has focused on developing fast algorithms for submodular maximization recently \citep{leskovec2007cost,badanidiyuru2014fast,mirzasoleiman2016fast,ene2018towards,mirzasoleiman2015lazier}. \citet{mirzasoleiman2015lazier} propose the first linear-time algorithm that achieves  a $1-1/e-\epsilon$ approximation, using only $O(n\log \frac{1}{\epsilon})$ value oracle queries. Their algorithm performs $k$ rounds: at each round, it draws a small random sample of items, and selects the item with the largest marginal utility  from the random sample. They show that their algorithm can achieve linear time complexity by carefully choosing the size of the random sample. Recently,
\citet{buchbinder2017comparing} extends the previous studies  to  non-monotone submodular maximization and they develop the first linear-time algorithm for this problem under non-adaptive setting.

 Our study focuses on the adaptive submodular maximization problem,  a stochastic variant of the classical non-adaptive submodular maximization problem.  The input of our problem is a set of items, each item  is in a particular state drawn from a known prior distribution. One must select an item before observing its actual state. An adaptive policy specifies which item to pick next based on the current observation. One classical example is called sensor selection, where we would like to select  a collection of $k$ sensors from $n$ candidate sensors to monitor some targets \citep{asadpour2016maximizing}. Each sensor has two possible states  failure or normal, whose realization is drawn from a known prior distribution. The only way to observe a sensor's state is to select that sensor. Assume that each sensor can monitor a known set of targets if its state is normal, otherwise,  if a sensor's state is failure, it can not monitor any targets. We would like to adaptively select a group of $k$ sensors that maximizes the total number of targets that can be monitored. A typical adaptive selection policy can be roughly described as follows: We first select some sensor and observe its state, then selects the next sensor by taking into account the observed state of the previously selected sensor, and this process repeats until we have selected $k$ sensors. To capture the adaptability of the selection process, \citet{golovin2011adaptive}  generalize the classical notion of submodularity and monotonicity by introducing \emph{adaptive submodularity}, whose formal definition is listed in Definition \ref{def:11}, and \emph{adaptive monotonicity}, whose formal definition is listed in Definition \ref{def:23}. They show that  a simple adaptive greedy algorithm achieves a $1-1/e$ approximation for maximizing a monotone adaptive submodular function subject to a cardinality constraint. Their algorithm requires $O(n\times k)$ value oracle queries. 
 While the literature on adaptive submodular maximization  \citep{chen2013near,tang2020influence,tang2020price,yuan2017adaptive,fujii2019beyond,gabillon2013adaptive,golovin2010near} typically assumes adaptive monotonicity, the more general problem of non-monotone adaptive maximization is first studied in \citep{gotovos2015non}. They propose an adaptive random greedy algorithm that achieves a $1/e$ approximation ratio. However, their result relies on the assumption that the objective function is adaptive submodular and \emph{pointwise submodular}, whose formal definition is listed in Definition \ref{def:121}. 
 Note that adaptive submodularity does not imply pointwise submodular \citep{guillory2010interactive,golovin2011adaptive}, it is not clear whether the same guarantee  holds without resorting to pointwise submodularity  or not. Moreover, their algorithm requires $O(n\times k)$ value oracle queries. Very recently, \citep{amanatidis2020fast} develop a constant approximate solution for maximizing a non-monotone adaptive submodular and pointwise submodular function subject to a knapsack constraint.

\textbf{ Our contributions.} Our first contribution is to present an enhanced analytical result by  showing that the adaptive random greedy algorithm achieves a $1/e$ approximation ratio under  adaptive submodularity (without resorting to pointwise submodularity). Our second contribution is to propose the first linear-time  algorithm  for the non-monotone adaptive submodular maximization problem subject to a cardinality constraint. Our algorithm  achieves a $1/e-\epsilon$ approximation ratio, using only $O(n\epsilon^{-2}\log \epsilon^{-1})$ value oracle queries. For monotone case, our algorithm achieves a  $1-1/e-\epsilon$ approximation ratio. Our proposed approach generalizes the non-adaptive linear-time algorithm proposed in \citep{buchbinder2017comparing} to the adaptive setting. At last,  for the monotone case, we propose a faster algorithm that achieves a $1-1/e-\epsilon$ approximation ratio in expectation  with $O(n \log  \frac{1}{\epsilon})$ value oracle queries. We also consider a more general constraint called partition matroid constraint, and develop a linear-time algorithm for monotone and fully adaptive submodular functions.

\section{Preliminaries}
In the rest of this paper, we use  $[m]$ to denote the set $\{1, 2, \cdots, m\}$, and use $|S|$ to denote the cardinality of a set $S$.

\subsection{Items and  States} The input of our problem is set  $E$ of $n$ items. Each item $e\in E$ has a particular state  from $O$.  We use $\phi: E\rightarrow O$ to denote a \emph{realization} and let $U$ denote the set of all realizations.   Let $\Phi=\{\Phi_e \mid e\in E\}$ denote a random realization drawn from a known prior probability distribution $p(\phi)=\{\Pr[\Phi=\phi]: \phi\in U\}$, where $\Phi_e \in O$ denotes a random realization of $e$. One must select an item before observing its state.  After selecting a subset of items, we observe the \emph{partial realization} $\psi$ of  those items. The \emph{domain} of $\psi$, which is denoted by $\mathrm{dom}(\psi)$, is defined  as the subset of items involved in $\psi$. For any realization $\phi$ and any  partial realization $\psi$, we say $\psi$ is consistent with $\phi$ if they are equal everywhere in $\mathrm{dom}(\psi)$. In this case, we write $\phi \sim \psi$. We say that $\psi$  is a \emph{subrealization} of  $\psi'$ if $\mathrm{dom}(\psi) \subseteq \mathrm{dom}(\psi')$ and they are equal everywhere in $\mathrm{dom}(\psi)$. In this case, we write  $\psi \subseteq \psi'$. We use $p(\phi\mid \psi)$ to denote the conditional distribution over realizations conditioned on  a partial realization $\psi$: $p(\phi\mid \psi) =\Pr[\Phi=\phi\mid \Phi\sim \psi ]$. There is a utility function $f$ from a subset of items and their states to a non-negative real number: $f: 2^{E}\times O^E\rightarrow \mathbb{R}_{\geq0}$.

\subsection{Policies and Problem Formulation} Based on the above notations, we can encode any adaptive policy  using a function $\pi$  from a set of partial realizations to a distribution of $E$, .e.g,  $\pi$ takes as input the partial realizations of selected items and outputs which item to select next.

\begin{definition}[Policy  Concatenation]
Given two policies $\pi$ and $\pi'$,  we define  $\pi @\pi'$ as a policy that runs $\pi$ first, and then runs $\pi'$. Note that running $\pi'$  does not rely on the observation obtained from running $\pi$.
\end{definition}

The expected  utility $f_{avg}(\pi)$ of a policy $\pi$ is
\begin{equation}
f_{avg}(\pi)=\mathbb{E}_{\Phi\sim p(\phi)}f(E(\pi, \Phi), \Phi)
\end{equation}
where $E(\pi, \phi)$ denotes the subset of items selected by $\pi$ under realization $\phi$.  We say a policy $\pi$ is feasible if it selects at most $k$ items for all realizations, that is, $|E(\pi, \Phi)|\leq k$ for all $\phi\in U$.

Our goal is to find a feasible policy  $\pi^{opt}$ that maximizes the expected utility:
\[\pi^{opt} \in \arg\max_{\pi \in \Omega} f_{avg}(\pi)\]
where $\Omega$ denotes the set of all feasible policies.

\subsection{Adaptive Submodularity and Monotonicity}
We next introduce the concept of adaptive submodularity. We start by introducing two notations.
\begin{definition}[Conditional Expected Marginal Utility of a set of Items]
\label{def:1}
Given any partial realization $\psi$ and any set of items $S$, the conditional expected marginal utility $\Delta(e \mid \psi)$ of $S$ conditioned on $\psi$ is
\[\Delta(S \mid \psi)=\mathbb{E}_{\Phi}[f(\mathrm{dom}(\psi)\cup S, \Phi)-f(\mathrm{dom}(\psi, \Phi))\mid \Phi \sim \psi]\]
where the expectation is taken over $\Phi$ with respect to $p(\phi\mid \psi)=\Pr(\Phi=\phi \mid \Phi \sim \psi)$.
\end{definition}

The algorithm is assumed to  access the objective function via a \emph{value oracle} that returns $\Delta(e \mid \psi)$ given any input element  $e\in E$ and partial realization $\psi$.

\begin{definition}[Conditional Expected Marginal Utility of a Policy]
\label{def:1}
Given any partial realization $\psi$ and a policy $\pi$, the conditional expected marginal utility $\Delta(\pi \mid \psi)$ of $\pi$ conditioned on $\psi$ is
\[\Delta(\pi\mid \psi)=\mathbb{E}_{\Phi}[f(\mathrm{dom}(\psi) \cup E(\pi, \Phi), \Phi)-f(\mathrm{dom}(\psi), \Phi)\mid \Phi\sim \psi]\]
where the expectation is taken over $\Phi$ with respect to $p(\phi\mid \psi)=\Pr(\Phi=\phi \mid \Phi \sim \psi)$.
\end{definition}

We next introduce the adaptive monotonicity and adaptive submodularity.

\begin{definition}\citep{golovin2011adaptive} [Adaptive Monotonicity]
\label{def:23}
A function $f: 2^E\times O^E\rightarrow \mathbb{R}_{\geq 0}$ is  adaptive monotone with respect to a prior distribution $ p(\phi)$, if for any partial realization $\psi$, it holds that
\[\Delta(e\mid \psi) \geq 0 \]
\end{definition}

\begin{definition}\citep{golovin2011adaptive}[Adaptive Submodularity]
\label{def:11}
A function $f: 2^E\times O^E\rightarrow \mathbb{R}_{\geq 0}$ is  adaptive submodular with respect to a prior distribution $ p(\phi)$, if for any two partial realizations $\psi$ and $\psi'$ such that $\psi\subseteq \psi'$, the following holds:
\[\Delta(e\mid \psi) \geq \Delta(e\mid \psi') \]
\end{definition}

For comparison purpose, we further introduce the \emph{pointwise submodularity}.
\begin{definition}\citep{golovin2011adaptive}[Pointwise Submodularity]
\label{def:121}
A function $f: 2^E\times O^E\rightarrow \mathbb{R}_{\geq 0}$  is  pointwise submodular if $f(S, \phi)$ is submodular in terms of $S\subseteq E$ for all $\phi\in U$. That is, for any $\phi\in U$, any two sets $S_1\subseteq E$ and $S_2 \subseteq E$ such that $S_1 \subseteq S_2$, and any item $e\notin S_2$, we have $f(S_1\cup\{e\}, \phi)- f(S_1, \phi) \geq f(S_2\cup\{e\}, \phi)- f(S_2, \phi)$.
\end{definition}

Note that adaptive submodularity does not  imply pointwise submodularity and vice versa.

%
%

\section{Revisiting Adaptive Random Greedy Policy}
We first revisit a simple random greedy policy, called Adaptive Random Greedy Policy $\pi^{arg}$, that is proposed in \citep{gotovos2015non}. \citet{gotovos2015non} show that $\pi^{arg}$ achieves a $1/e$ approximation ratio under the condition that $f$ is adaptive submodular and pointwise submodular. Because adaptive submodularity does not  imply pointwise submodularity and vice versa, it is not clear whether adaptive submodularity is a sufficient condition for achieving the above guarantee.  We provide a positive answer to this question by showing that $\pi^{arg}$ achieves a $1/e$ approximation ratio  under adaptive submodularity (without resorting to pointwise submodularity).

 We first explain the idea of $\pi^{arg}$ (Algorithm \ref{alg:LPP1}). We first add a set $D$ of $2k-1$ dummy items to the ground set, such that, for any $d \in D$, and any
partial realization $\psi$, we have $\Delta(d \mid \psi) =0$. Let $E'=E\cup D$. We add these dummy items to ensure that $\pi^{arg}$  never selects an item with negative marginal utility. Clearly, these dummy items can be safely removed from the solution returned from any policy without affecting its utility. $\pi^{arg}$ runs round by round: Starting with an empty set and  at each round $r\in[k]$, $\pi^{arg}$  randomly selects an
item from the set $M(\psi^{r-1})$, which contains the $k$ items with the largest marginal utility to the current partial realization $\psi^{r-1}$.

\begin{algorithm}[hptb]
\caption{Adaptive Random Greedy Policy $\pi^{arg}$}
\label{alg:LPP1}
\begin{algorithmic}[1]
\STATE $A=\emptyset; r=1$.
\WHILE {$r \leq k$}
\STATE observe $\psi^{r-1}$;
\STATE $M(\psi^{r-1})\leftarrow \arg\max_{M\subseteq E'; |M|\leq k} \sum_{e\in E'}\Delta(e\mid \psi^{r-1})$;
\STATE sample $e_r$ uniformly at random from $M(\psi^{r-1})$;
\STATE $A\leftarrow A\cup \{e_r\}$; $r\leftarrow r+1$;
\ENDWHILE
\RETURN $A$
\end{algorithmic}
\end{algorithm}


We next present the main theorem.
\begin{theorem}
\label{thm:1}
If $f$ is adaptive submodular, then the Adaptive Random Greedy Policy $\pi^{arg}$ achieves a $1/e$ approximation ratio in expectation  with $O(n k)$ value oracle queries.
\end{theorem}

\emph{Proof:} We first prove the time complexity of $\pi^{arg}$. As $\pi^{arg}$ performs $k$ rounds and each rounds performs $O(n)$ value oracle queries, the time complexity of $\pi^{arg}$ is $O(nk)$. We next prove the approximation ratio of $\pi^{arg}$.

For every $r\in [k]$, let $\pi^{arg}_r$ denote the policy that runs  $\pi^{arg}$ for $r$ rounds. We first provide a preparation lemma as follows.

\begin{lemma}
\label{lem:1111}When $f$ is adaptive submodular, for every $r\in [k]$,
\begin{equation*}
f_{avg}(\pi^{opt}@\pi^{arg}_r) \geq (1-\frac{1}{k})^r f_{avg} (\pi^{opt})
\end{equation*}
\end{lemma}
\emph{Proof:} {Fix $r\in [k]$ and a partial realization $\psi^{o@arg_{(r-1)}}= \psi^{opt}\cup\psi^{r-1}$ that is observed after running $\pi^{opt}@\pi^{arg}_{r-1}$, where $\psi^{opt}$ is the partial realization after running $\pi^{opt}$ and $\psi^{r-1}$ is the partial realization after running $\pi^{arg}_{r-1}$.} Observe that,
\begin{eqnarray}
\mathbb{E}_{e_r}[\Delta(e_r \mid\psi^{o@arg_{(r-1)}})]&=& \frac{1}{k} \sum_{e\in M(\psi^{r-1})} \Delta(e \mid\psi^{o@arg_{(r-1)}})~\nonumber\\
&\geq& \frac{1}{k}\Delta(M(\psi^{r-1}) \mid \psi^{o@arg_{(r-1)}})~\nonumber\\
&=& \frac{1}{k}(\mathbb{E}_{\Phi}[f_{avg}(\mathrm{dom}(\psi^{o@arg_{(r-1)}})\cup M(\psi^{r-1}), \Phi)\mid\Phi\sim \psi^{o@arg_{(r-1)}}] ~\nonumber\\ &&\quad\quad\quad\quad\quad\quad\quad\quad\quad-\mathbb{E}_{\Phi}[f(\mathrm{dom}(\psi^{o@arg_{(r-1)}}), \Phi)\mid \Phi\sim \psi^{o@arg_{(r-1)}}])~\nonumber\\
&\geq& -\frac{\mathbb{E}_{\Phi}[f(\mathrm{dom}(\psi^{o@arg_{(r-1)}}), \Phi)\mid \Phi\sim \psi^{o@arg_{(r-1)}}]}{k} \label{eq:ddd}
\end{eqnarray} where  the first inequality is due to $f$ is adaptive submodular. Unfixing $\psi^{o@arg_{(r-1)}}$, taking the expectation over {$(\Psi^{opt}, \Psi^{r-1})$}, we have
\begin{equation}
\label{eq:9999}
\mathbb{E}_{\Psi^{o@arg_{(r-1)}}}\left[\mathbb{E}_{e_r}[\Delta(e_r \mid\Psi^{o@arg_{(r-1)}})]\right] = f_{avg}(\pi^{opt}@\pi^{arg}_{r})-f_{avg}(\pi^{opt}@\pi^{arg}_{r-1}) \geq -\frac{f_{avg}(\pi^{opt}@\pi^{arg}_{r-1})}{k}
\end{equation}

We are now ready to prove this lemma by induction on $r$. When $r=1$, this lemma holds because $f_{avg}(\pi^{opt}@\pi^{arg}_0) = f_{avg}(\pi^{opt}) \geq (1-\frac{1}{k})^0f_{avg} (\pi^{opt})$. Assume this lemma holds for $r'<r$, we next prove it for $r>0$.
\begin{eqnarray*}
f_{avg}(\pi^{opt}@\pi^{arg}_r) &=& f_{avg}(\pi^{opt}@\pi^{arg}_{r-1})+ (f_{avg}(\pi^{opt}@\pi^{arg}_{r})-f_{avg}(\pi^{opt}@\pi^{arg}_{r-1}))\\
&\geq&  f_{avg}(\pi^{opt}@\pi^{arg}_{r-1})-\frac{f_{avg}(\pi^{opt}@\pi^{arg}_{r-1})}{k}\\
&=& (1-\frac{1}{k}) f_{avg}(\pi^{opt}@\pi^{arg}_{r-1})\\
&\geq & (1-\frac{1}{k}) (1-\frac{1}{k})^{r-1} f_{avg}(\pi^{opt})\\
&=& (1-\frac{1}{k})^{r} f_{avg}(\pi^{opt})
\end{eqnarray*} The first inequality is due to Eq. (\ref{eq:9999}) and the second inequality follows from the inductive assumption. $\Box$

Now we are ready to prove the theorem. 
The expectation $\mathbb{E}_{ \Psi^{r-1}}[\cdot]$ is taken over all partial realizations $\Psi^{r-1}$. Then we have
\begin{eqnarray}
f_{avg}(\pi^{arg}_r) - f_{avg}(\pi^{arg}_{r-1}) &=& \mathbb{E}_{ \Psi^{r-1}}[\mathbb{E}_{e_r}[\Delta(e_r \mid\Psi^{r-1})]]\nonumber\\
&=& \frac{1}{k} \mathbb{E}_{\Psi^{r-1}} [\sum_{e\in M(\Psi^{r-1})} \Delta(e \mid\Psi^{r-1})]\nonumber\\
&\geq& \frac{1}{k} \mathbb{E}_{ \Psi^{r-1}} [ \Delta(\pi^{opt} \mid\Psi^{r-1})]\nonumber\\
&=& \frac{f_{avg}(\pi^{opt}@\pi^{arg}_{r-1})-  f_{avg}(\pi^{arg}_{r-1})}{k}\nonumber\\
&\geq& \frac{(1-\frac{1}{k})^{r-1} f_{avg}(\pi^{opt})-  f_{avg}(\pi^{arg}_{r-1})}{k}\label{eq:bbbb}
\end{eqnarray}
The second equality is due to the design of $\pi^{arg}$, the first inequality is due to $f$ is adaptive submodular and Lemma 1 in \citep{gotovos2015non}, and the second inequality is due to Lemma \ref{lem:1111}.

 We next prove
 \begin{equation}\label{eq:xxx}
 f_{avg}(\pi^{arg}_{r})\geq \frac{r}{k} (1-\frac{1}{k})^{r-1}f_{avg}(\pi^{opt})
 \end{equation} by induction on $r$. For $r=0$,  $f_{avg}(\pi^{arg}_{0})\geq 0 \geq \frac{0}{k} (1-\frac{1}{k})^{0-1}f_{avg}(\pi^{opt})$. Assume Eq. (\ref{eq:xxx}) is true for $r'<r$, let us
prove it for $r$.
\begin{eqnarray*}
 f_{avg}(\pi^{arg}_{r}) &\geq&  f_{avg}(\pi^{arg}_{r-1})+ \frac{(1-\frac{1}{k})^{r-1} f_{avg}(\pi^{opt})-  f_{avg}(\pi^{arg}_{r-1})}{k}\\
 &=& (1-1/k)f_{avg}(\pi^{arg}_{r-1})+\frac{(1-\frac{1}{k})^{r-1} f_{avg} (\pi^{opt})}{k}\\
  &\geq& (1-1/k)\cdot ((r-1)/k)\cdot (1-1/k)^{r-2}\cdot f_{avg}(\pi^{opt})+\frac{(1-\frac{1}{k})^{r-1} f_{avg} (\pi^{opt})}{k}\\
  &\geq&  (r/k)\cdot (1-1/k)^{r-1}\cdot f_{avg}(\pi^{opt})
\end{eqnarray*}
The first equality is due to Eq. (\ref{eq:bbbb}), the second inequality is due to the inductive assumption.
When $r=k$, we have $ f_{avg}(\pi^{arg}_{r}) \geq  (1-1/k)^{k-1}\cdot f_{avg}(\pi^{opt}) \geq (1/e) f_{avg}(\pi^{opt})$. This finishes the proof of the theorem. $\Box$

For monotone case, \citet{gotovos2015non} show that $\pi^{arg}$  achieves a $1-1/e$ approximation ratio.
\begin{theorem}\citep{gotovos2015non}
\label{thm:4}
If $f$ is adaptive submodular and adaptive monotone, then the Adaptive Random Greedy Policy $\pi^{arg}$ achieves a $1-1/e$ approximation ratio in expectation  with $O(n k)$ value oracle queries.
\end{theorem}

\section{Linear-Time Adaptive Policy}
We next present the Linear-Time Adaptive Policy $\pi^{lt}$ (Algorithm \ref{alg:LPP2}). Later we show that it achieves a $1/e-\epsilon$ approximation ratio, using only $O(n\epsilon^{-2}\log \epsilon^{-1})$ value oracle queries.

We first explain the idea of $\pi^{lt}$ which generalizes the non-adaptive linear-time algorithm \citep{buchbinder2017comparing} to the adaptive setting.  $\pi^{lt}$ has two parameters $q$ and $s$ and it works for $k$ rounds: Starting with an empty set. At each round $r\in[k]$, $\pi^{lt}$ first samples a random set $S_r$ of size $\min\{\lceil qn \rceil, n\}$ from $E$. Let $e'_r$ be the item of $S_r$ that has the $\lceil d_r \rceil$-th largest marginal utility to the current partial realization $\psi^{r-1}$ (by abuse of notation), where $d_r$ is a random value  sampled from range $(0,s]$. $\pi^{lt}$  adds $e'_r$ to the solution if $\Delta(e'_r\mid \psi^{r-1})\geq 0$.

\begin{algorithm}[hptb]
\caption{Linear-Time Adaptive Policy $\pi^{lt}$}
\label{alg:LPP2}
\begin{algorithmic}[1]
\STATE $A=\emptyset; r=1$.
\WHILE {$r \leq k$}
\STATE observe $\psi^{r-1}$;
\STATE Let $S_r$ be a random set of size $\min\{\lceil qn \rceil, n\}$ sampled uniformly at random  from $E$;
\STATE Let $d_r$ be a random value  sampled from range $(0,s]$;
\STATE Let $e'_r$ be the item of $S_r$ that has the $\lceil d_r \rceil$-th largest marginal utility to $\psi^{r-1}$;
\IF {$\Delta(e'_r\mid \psi^{r-1})\geq 0$}
\STATE $e_r \leftarrow e'_r$
\ELSE
\STATE $e_r \leftarrow \emptyset$
\ENDIF
\STATE  $A\leftarrow A\cup \{e_r\}$;
\STATE  $r\leftarrow r+1$;
\ENDWHILE
\RETURN $A$
\end{algorithmic}
\end{algorithm}

By setting $q=8k^{-1}\epsilon^{-2}\cdot\ln (2\epsilon)^{-1}$ and $s=k\min\{\lceil qn \rceil, n\}/n$, we have the following theorem.
\begin{theorem}
\label{thm:2}
If $f$ is adaptive submodular, then the Linear-Time Adaptive Policy $\pi^{lt}$ achieves a $(1/e-\epsilon)$ approximation ratio in expectation  with $O(n\epsilon^{-2}\log \epsilon^{-1})$ value oracle queries.
\end{theorem}
\emph{Proof:} We first prove the time complexity of  $\pi^{lt}$. Observe that  $\pi^{lt}$ performs $k$ rounds, and each rounds performs $\min\{\lceil qn \rceil, n\}$ value oracle queries. Thus the total number of value oracle queries is bounded by $\min\{\lceil qn \rceil, n\}\times k\leq (qn+1)\times k$. Since we set $q=8k^{-1}\epsilon^{-2}\cdot\ln (2\epsilon)^{-1}$, we have $\min\{\lceil qn \rceil, n\}\times k \leq (qn+1)\times k = O(n\epsilon^{-2}\log \epsilon^{-1})$.

We next prove the approximation ratio of $\pi^{lt}$. Note that when $\min\{\lceil qn \rceil, n\}=n$, we have $s=k$, and   $\pi^{lt}$ is reduced to  $\pi^{arg}$. It follows that $\pi^{lt}$ achieves a $1/e$ approximation ratio due to Theorem \ref{thm:1}. In the rest of the proof, we assume that $\lceil qn \rceil< n$.

We first provide two preparation lemmas: Lemma \ref{lem:a} and Lemma \ref{lem:2222}.  For any $r\in [k]$, let $\pi^{lt}_{r}$ denote the policy that runs $\pi^{lt}$ for $r$ rounds.
\begin{lemma}
\label{lem:a}When $f$ is adaptive submodular, for every $r\in [k]$,
\begin{equation}
f_{avg}(\pi^{opt}@\pi^{lt}_r) \geq (1-\frac{1}{k})^r f_{avg} (\pi^{opt})
\end{equation}
\end{lemma}
\emph{Proof:} Fix $r\in [k]$ and a partial realization {$\psi^{o@lt_{(r-1)}}=\psi^{opt}\cup\psi^{r-1}$} that is observed after running $\pi^{opt}@\pi^{lt}_{r-1}$. Observe that,
\begin{eqnarray}
\mathbb{E}_{e_r}[\Delta(e_r \mid\psi^{o@{lt}_{(r-1)}})]&=&\sum_{e\in E\cup\{\emptyset\}} \Pr[e \mbox{ is selected at round $r$}]  \Delta(e \mid\psi^{o@{lt}_{(r-1)}})\nonumber\\
&=&\sum_{e\in E} \Pr[e \mbox{ is selected at round $r$}]  \Delta(e \mid\psi^{o@{lt}_{(r-1)}})\nonumber\\
&\geq&\sum_{e\in E\setminus E^+} \Pr[e \mbox{ is selected at round $r$}]  \Delta(e \mid\psi^{o@{lt}_{(r-1)}})\nonumber\\
&\geq& \sum_{e\in  E\setminus E^+} \frac{\lceil qn \rceil /n}{s} \Delta(e \mid\psi^{o@{lt}_{(r-1)}})\nonumber\\
&\geq& \frac{1}{k} \sum_{e\in E\setminus E^+} \Delta(e \mid\psi^{o@{lt}_{(r-1)}})\nonumber\\
&\geq& \frac{1}{k}\Delta(E\setminus E^+ \mid \psi^{o@{lt}_{(r-1)}})\nonumber\\
&=& \frac{1}{k}(\mathbb{E}_{\Phi}[f_{avg}(\mathrm{dom}(\psi^{o@{lt}_{(r-1)}}
)\cup (E\setminus E^+), \Phi)\mid\Phi\sim \psi^{o@{lt}_{(r-1)}}]\nonumber\\
&&\quad\quad\quad\quad\quad\quad\quad\quad-\mathbb{E}_{\Phi}[f(\mathrm{dom}(\psi^{o@{lt}_{(r-1)}}), \Phi)\mid \Phi\sim \psi^{o@{lt}_{(r-1)}}])\nonumber\\
&\geq& -\frac{\mathbb{E}_{\Phi}[f(\mathrm{dom}(\psi^{o@{lt}_{(r-1)}}), \Phi)\mid \Phi\sim \psi^{o@{lt}_{(r-1)}}]}{k} \label{eq:d}
\end{eqnarray} where $E^+=\{e\mid e\in E; \Delta(e \mid\psi^{o@{lt}_{(r-1)}})\geq 0\}$. The second equality is due to $\Delta(\emptyset \mid\psi^{o@{lt}_{(r-1)}})=0$, the first inequality is due to the definition of $E^+$, the second inequality is due to $\Pr[e \mbox{ is selected at round $r$}]\leq \frac{\lceil qn \rceil /n}{s}$ for all $e\in E$ and $\Delta(e \mid\psi^{o@{lt}_{(r-1)}})< 0$ for all $e\in E\setminus E^+$, the third inequality is due to $q=8k^{-1}\epsilon^{-2}\cdot\ln (2\epsilon)^{-1}$ and $s=k\lceil qn \rceil/n$, the forth inequality is due to $f$ is adaptive submodular.

Unfixing $\psi^{o@lt_{(r-1)}}$, taking the expectation over {$(\Psi^{opt}, \Psi^{r-1})$}, we have
\begin{equation}
\label{eq:999}
\mathbb{E}_{\Psi^{o@lt_{(r-1)}}}\left[\mathbb{E}_{e_r}[\Delta(e_r \mid\Psi^{o@lt_{(r-1)}})]\right] = f_{avg}(\pi^{opt}@\pi^{lt}_{r})-f_{avg}(\pi^{opt}@\pi^{lt}_{r-1}) \geq -\frac{f_{avg}(\pi^{opt}@\pi^{lt}_{r-1})}{k}
\end{equation}
The rest of the proof is similar to the proof of Lemma \ref{lem:1111}, thus omitted here. $\Box$

The proof of the following lemma is provided in appendix.
\begin{lemma}
\label{lem:2222}
When $f$ is adaptive submodular, for every $r\in [k]$, $f_{avg}(\pi^{lt}_{r})-f_{avg}(\pi^{lt}_{r-1})\geq (1-\epsilon)\frac{f_{avg}(\pi^{opt}@\pi^{lt}_{r-1})-f_{avg}(\pi^{lt}_{r-1})}{k}$.
\end{lemma}

Lemma \ref{lem:a} and Lemma \ref{lem:2222} imply that
 \begin{eqnarray}
 f_{avg}(\pi^{lt}_{r})-f_{avg}(\pi^{lt}_{r-1})&\geq& (1-\epsilon)\frac{(1-\frac{1}{k})^{r-1} f_{avg} (\pi^{opt})-f_{avg}(\pi^{lt}_{r-1})}{k}\\
 &\geq& \frac{[(1-\frac{1}{k})^{r-1}-\epsilon] f_{avg} (\pi^{opt})-f_{avg}(\pi^{lt}_{r-1})}{k} \label{eq:zzz}
 \end{eqnarray}

 To prove this theorem, it suffice to show that
 \begin{equation}
 \label{eq:aaa}
 f_{avg}(\pi^{lt}_{r})\geq (r/k)\cdot [(1-1/k)^{r-1}-\epsilon]\cdot f_{avg}(\pi^{opt})
 \end{equation}
  for all $r\in [k]$.
 Similar to \citep{buchbinder2017comparing} (the proof of Theorem 4.2), we prove Eq. (\ref{eq:aaa}) by induction on $r$. For $r=0$,  $f_{avg}(\pi^{lt}_{0})\geq 0 \geq (0/k)\cdot [(1-1/k)^{0-1}-\epsilon]\cdot f_{avg}(\pi^{opt})$. Assume Eq. (\ref{eq:aaa}) is true for $r'<r$, let us
prove it for $r$.
\begin{eqnarray*}
 f_{avg}(\pi^{lt}_{r}) &\geq&  f_{avg}(\pi^{lt}_{r-1})+ \frac{[(1-\frac{1}{k})^{r-1}-\epsilon] f_{avg} (\pi^{opt})-f_{avg}(\pi^{lt}_{r-1})}{k}\\
 &=& (1-1/k)f_{avg}(\pi^{lt}_{r-1})+\frac{[(1-\frac{1}{k})^{r-1}-\epsilon] f_{avg} (\pi^{opt})}{k}\\
  &\geq& (1-1/k)\cdot ((r-1)/k)\cdot [(1-1/k)^{r-2}-\epsilon]\cdot f_{avg}(\pi^{opt})+\frac{[(1-\frac{1}{k})^{r-1}-\epsilon] f_{avg} (\pi^{opt})}{k}\\
  &\geq&  (r/k)\cdot [(1-1/k)^{r-1}-\epsilon]\cdot f_{avg}(\pi^{opt})
\end{eqnarray*}
The first inequality is due to (\ref{eq:zzz}), the second inequality is due to the inductive assumption. When $r=k$, we have $ f_{avg}(\pi^{lt}_{r}) \geq  [(1-1/k)^{k-1}-\epsilon]\cdot f_{avg}(\pi^{opt}) \geq (1/e-\epsilon) f_{avg}(\pi^{opt})$. This finishes the proof of the theorem. $\Box$

\paragraph{Performance bound for monotone case.} We next show that if $f$ is adaptive monotone, the performance bound of $\pi^{lt}$ is improved to $1-1/e-\epsilon$.

\begin{theorem}
\label{thm:3}
If $f$ is adaptive submodular and adaptive monotone, then the Linear-Time Adaptive Policy $\pi^{lt}$ achieves a $1-1/e-\epsilon$ approximation ratio in expectation  with $O(n\epsilon^{-2}\log \epsilon^{-1})$ value oracle queries.
\end{theorem}
\emph{Proof:} The time complexity result inherits from Theorem \ref{thm:2}. We next prove the performance bound of $\pi^{lt}$.  When $f$ is adaptive submodular and adaptive monotone, Lemma \ref{lem:2222} still holds. Thus, $f_{avg}(\pi^{lt}_{r})-f_{avg}(\pi^{lt}_{r-1})\geq (1-\epsilon)\frac{f_{avg}(\pi^{opt}@\pi^{lt}_{r-1})-f_{avg}(\pi^{lt}_{r-1})}{k}\geq (1-\epsilon)\frac{f_{avg}(\pi^{opt})-f_{avg}(\pi^{lt}_{r-1})}{k}$, where the second inequality is due to $f$ is adaptive monotone. It follows that $f_{avg}(\pi^{lt}_{k})\geq (1-(1-\frac{1-\epsilon}{k})^k)f_{avg}(\pi^{opt})\geq (1-(1/e)^{1-\epsilon})f_{avg}(\pi^{opt})\geq (1-1/e-\epsilon)f_{avg}(\pi^{opt})$ through induction on $r$. $\Box$
\section{Monotone Case: A Faster Algorithm}
We next propose a faster algorithm \emph{Adaptive Stochastic Greedy}, denoted by  $\pi^{asg}$, for maximizing a monotone adaptive submodularity function. As compared with $\pi^{lt}$ which needs $O(n\epsilon^{-2}\log \epsilon^{-1})$ value oracle queries, our enhanced policy $\pi^{asg}$ achieves the same performance guarantee with $O(n \log  \frac{1}{\epsilon})$ value oracle queries. 
The details of our algorithm are listed in Algorithm \ref{alg:LPP13}. Similar to the classic adaptive greedy algorithm, $\pi^{asg}$ runs round by round: Starting with an empty set and  at each round, it selects one item that maximizes the expected marginal utility based on the current observation. What is different from the adaptive greedy algorithm, however, is that at each round $r\in[k]$, $\pi^{asg}$ first samples a set $S_r$ of size $\frac{n}{k}\log\frac{1}{\epsilon}$ uniformly at random and then selects the item with the largest conditional expected marginal utility from $S_r$. Our approach is a natural extension of the \emph{Stochastic Greedy} algorithm \citep{mirzasoleiman2015lazier}, the first linear-time algorithm for maximizing a monotone submodular function under the non-adaptive setting, we generalize their results to the adaptive setting. We will show that $\pi^{asg}$  achieves $1-1/e-\epsilon$ approximation ratio for maximizing an adaptive monotone and submodular function, and it has linear running time independent of the cardinality constraint $k$. It was worth noting that the technique of lazy updates \citep{minoux1978accelerated}  can be used to further accelerate the computation of our algorithms in practice.

\begin{algorithm}[hptb]
\caption{Adaptive Stochastic Greedy $\pi^{asg}$}
\label{alg:LPP13}
\begin{algorithmic}[1]
\STATE $A=\emptyset; r=1$.
\WHILE {$r \leq k$}
\STATE observe $\psi^{r-1}$;
\STATE $S_r\leftarrow$ a random set sampled uniformly at random  from $E$;
\STATE $e_r\leftarrow \arg\max_{e \in S_r}\Delta(e\mid \psi^{r-1})$;
\STATE $A\leftarrow A\cup \{e_r\}$; $r\leftarrow r+1$;
\ENDWHILE
\RETURN $A$
\end{algorithmic}
\end{algorithm}

We next present the main theorem.
\begin{theorem}
\label{thm:main}
If $f$ is adaptive submodular and adaptive monotone, then the Adaptive Stochastic Greedy policy $\pi^{asg}$ achieves a $1-1/e-\epsilon$ approximation ratio in expectation  with $O(n \log  \frac{1}{\epsilon})$ value oracle queries.
\end{theorem}

\emph{Proof:} To prove this theorem, we follow an argument similar to the one from \citep{mirzasoleiman2015lazier} for the proof of Theorem 1, but extended to the adaptive setting. We first prove the time complexity of $\pi^{asg}$. Recall that we set the size of $S_r$ to $\frac{n}{k}\log\frac{1}{\epsilon}$, thus the total number of value oracle queries is at most $k\times \frac{n}{k}\log\frac{1}{\epsilon}=n \log  \frac{1}{\epsilon}$.

 Before proving the approximation ratio of $\pi^{asg}$, we first provide a preparation lemma as follows.
\begin{lemma}
\label{lem:aX}
Given any partial realization $\psi$, let $A^*\in \arg\max_{A\subseteq  E, |A|= k} \sum_{e\in A} \Delta(e\mid \psi)$ denote the $k$ largest items in terms of the expected marginal utility conditioned on having observed $\psi$. Assume $R$ is sampled  uniformly at random  from $E$, and the size of $R$ is $|R|=\frac{n}{k}\log\frac{1}{\epsilon}$. We have $\Pr[R\cap  A^*\neq \emptyset]\geq 1-\epsilon$.
\end{lemma}

The proof of Lemma \ref{lem:aX} is moved to Appendix. Given Lemma \ref{lem:aX} in hand, now we are ready to bound the approximation ratio of $\pi^{asg}$. Let $A_r=\{e_1, e_2, \cdots, e_r\}$ denote the first $r$ items selected by  $\pi^{asg}$, and we still use $\psi^{r-1}$ to denote the partial realization observed before selecting the $r$-th item, e.g., $\mathrm{dom}(\psi^{r-1})=A_{r-1}$. Our goal is to estimate the increased utility  $f_{avg}(\pi^{asg}_{r})-f_{avg}(\pi^{asg}_{r-1})$ during one round of our algorithm where $\pi^{asg}_r$ denotes a policy that runs $\pi^{asg}$ until it selects $r$ items.
Given a partial realization $\psi^{r-1}$  (by abuse of notation) after running $\pi^{asg}_{r-1}$, denote by $A_r^*\in \arg\max_{A\subseteq  E, |A|= k} \sum_{e\in A} \Delta(e\mid \psi^{r-1})$ the $k$ largest items in terms of the expected marginal utility conditioned on having observed $\psi^{r-1}$. Recall that  $S_r$ is sampled uniformly at random from $E$, each item in $A_r^*$ is equally likely to be contained in $S_r$. Moreover, the size of $S_r$ is set to $\frac{n}{k}\log\frac{1}{\epsilon}$. It follows that
\begin{eqnarray*}
\mathbb{E}_{e_r}[\Delta(e_r\mid \psi^{r-1})]&=&\frac{1}{k}\Pr[S_r\cap  A_r^* \neq \emptyset] \sum_{e\in A_r^*} \Delta(e\mid \psi^{r-1})\\
&\geq&\frac{1}{k}(1-\epsilon)\sum_{e\in A_r^*} \Delta(e\mid \psi^{r-1})
\end{eqnarray*} where the inequality is due to Lemma \ref{lem:aX}. Then we have
\[\mathbb{E}_{e_r}[\Delta(e_r\mid \psi^{r-1})] \geq (1-\epsilon) \frac{1}{k}\sum_{e\in A^*_r} \Delta(e\mid \psi^{r-1})\geq (1-\epsilon) \frac{1}{k} \Delta(\pi^{opt}\mid \psi^{r-1})\] where the second inequality is due to Lemma 6 in \citep{golovin2011adaptive}.
Therefore
\[\mathbb{E}_{e_r}[f_{avg}(\pi^{asg}_{r+1})-f_{avg}(\pi^{asg}_r)\mid \psi^{r-1}] = \mathbb{E}_{e_r}[\Delta(e_r\mid \psi^{r-1})] \geq (1-\epsilon) \frac{1}{k} \Delta(\pi^{opt}\mid \psi^{r-1})\]

 By abuse of notation, let $\Psi^{r-1}$ denote a random partial realization after running $\pi^{asg}_{r-1}$, then we have
\begin{eqnarray}
f_{avg}(\pi^{asg}_{r})-f_{avg}(\pi^{asg}_{r-1})=\mathbb{E}_{\Psi_{r-1}}\left[\mathbb{E}_{e_r}[\Delta(e_r\mid \Psi_{r-1})]\mid \Psi^{r-1}\right]  &\geq& (1-\epsilon) \frac{1}{k} \mathbb{E}_{ \Psi_r}[ \Delta(\pi^{opt}\mid \Psi^{r-1})]\nonumber\\
&=& (1-\epsilon) \frac{1}{k} ( f_{avg}(\pi^{asg}_{r-1}@\pi^{opt})-f_{avg}(\pi^{asg}_{r-1}))\nonumber\\
&\geq&(1-\epsilon)\frac{1}{k} (f_{avg}(\pi^{opt})-f_{avg}(\pi^{asg}_{r-1}))\label{eq:c}
\end{eqnarray}
Note that the first expectation is taken over two sources of randomness: one source is the randomness in choosing $e_r$, and the other source is the randomness in the partial realization $\Psi^{r-1}$. The second inequality is due to $f$ is adaptive monotone.  Based on (\ref{eq:c}), we have $f_{avg}(\pi^{asg})=f_{avg}(\pi^{asg}_k)\geq (1-1/e-\epsilon)f_{avg}(\pi^{opt})$ using induction. $\Box$

\section{Monotone Case: Extension to Partition Matroid Constraint}
We next study the monotone adaptive submodular maximization problem subject to a partition matroid constraint.

Let $B_1, B_2, \ldots, B_b$ be a collection of disjoint subsets
of $E$. Given a set of $b$ integers $\{d_i\mid i\in [b]\}$, define  $\Omega=\{\pi\mid  \forall \phi, \forall i\in[b], |E(\pi, \phi) \cap B_i|\leq d_i\}$ as the set of all feasible policies. When the objective function is adaptive monotone and adaptive submodular, we develop a $1/2$ approximate solution using $O(\sum_{i\in[b]} d_i |B_i|)$ value oracle queries. Then we present a linear-time adaptive policy whose runtime is independent of $d_i$. However, our analysis requires stronger assumptions about the objective function, e.g., we assume that the objective function is  fully adaptive submodular, which is a stricter condition than the adaptive submodularity.

\subsection{Locally Greedy Policy}
 Before presenting our linear-time algorithm, we first introduce a \emph{Locally Greedy} policy $\pi^{local}$. The basic idea of $\pi^{local}$ is as follows: Starting with an empty set, $\pi^{local}$ first selects $d_1$ number of items from $B_1$ in a greedy manner, i.e., iteratively adds items that maximize the conditional expected marginal utility conditioned on the realizations of  already selected items; $\pi^{local}$ then selects $d_2$ number of items from $B_2$ in the same greedy manner, and so on. Note that  $\pi^{local}$ does not rely on any specific ordering of $B_i$, and this motivates the term ``locally greedy'' that we use to name this policy. We can view $\pi^{local}$ as an adaptive version of the locally greedy algorithm proposed in  \citep{fisher1978analysis}.

 We first introduce some additional concepts.  Given a policy $\pi$, for any $i\in [b]$ and $j\in [d_i]$, we define 1) its \emph{level-$(i,j)$-truncation} $\pi_{ij}$ as a policy that runs $\pi$ until it selects  $j$ items from $B_i$ and 2) its \emph{strict level-$(i,j)$-truncation} $\pi_{\leftarrow ij}$ as a policy that runs $\pi$ until right before it selects  the $j$-th item  from $B_i$. It is clear that $f_{avg}(\pi)=\sum_{i\in[b], j\in[d_i]}(f_{avg}(\pi_{ij})-f_{avg}(\pi_{\leftarrow ij}))$.

We next show that if $f$ is adaptive submodular and adaptive monotone, the expected utility of $\pi^{local}$ is at least half of the optimal solution.
\begin{lemma}
\label{lem:6}
If $f$ is adaptive submodular and adaptive monotone, then  $f_{avg}(\pi^{local}) \geq  f_{avg}(\pi^{opt})/2$.
\end{lemma}

\emph{Proof:} Consider a policy $\pi^{local} @\pi^{opt}_{\leftarrow ij}$ that runs {$\pi^{local}$} first, then runs the strict level-$(i,j)$-truncation of $\pi^{opt}$. Let {$\psi^{l@o_{\leftarrow ij}}=\psi^{local}\cup \psi^{opt}_{\leftarrow ij}$} denote the partial realization obtained after running $\pi^{local} @\pi^{opt}_{\leftarrow ij}$. Based on this notation, we use $\psi^{l@o_{\leftarrow ij}}_{l_{\leftarrow ij }} \subseteq \psi^{l@o_{\leftarrow ij}}$ to denote a subrealization of $\psi^{l@o_{\leftarrow ij}}$ obtained after running the strict level-$(i,j)$-truncation of $\pi^{local}$. Conditioned on $\psi^{l@o_{\leftarrow ij}}$, assume $\pi^{local}$ selects $e^{local} _{ij}$ as the $j$-th item from $B_i$, and $\pi^{opt}$ selects $e^{opt}_{ij}$ as the $j$-th item from $B_i$. We first bound the expected marginal utility $\Delta(e^{local} _{ij}\mid \psi^{l@o_{\leftarrow ij}}_{l_{\leftarrow ij }})$ of $e^{local} _{ij}$ conditioned on having observed $\psi^{l@o_{\leftarrow ij}}_{l_{\leftarrow ij }}$.
\begin{eqnarray}
\Delta(e^{local} _{ij}\mid \psi^{l@o_{\leftarrow ij}}_{l_{\leftarrow ij }})&=& \max_{e\in B_i}\Delta(e \mid \psi^{l@o_{\leftarrow ij}}_{l_{\leftarrow ij }})\nonumber\\
&\geq& \max_{e\in B_i}\Delta(e \mid \psi^{l@o_{\leftarrow ij}})\nonumber\\
&\geq& \Delta(e^{opt}_{ij} \mid \psi^{l@o_{\leftarrow ij}})\nonumber\\
&=& f_{avg}(\pi^{local} @\pi^{opt}_{ij}\mid \psi^{l@o_{\leftarrow ij}})-f_{avg}(\pi^{local} @\pi^{opt}_{\leftarrow ij}\mid \psi^{l@o_{\leftarrow ij}}) \label{eq:ddd1}
\end{eqnarray}
The first equality is due to $\pi^{local}$ selects an item that maximizes the conditional expected marginal utility. The first inequality is due to the assumption that $f$ is adaptive submodular.

Let $\Psi^{l@o_{\leftarrow ij}}$ denote a random partial realization obtained after running $\pi^{local} @\pi^{opt}_{\leftarrow ij}$. Taking the expectation of $\Delta(e^{local} _{ij}\mid \Psi^{l@o_{\leftarrow ij}}_{l_{\leftarrow ij }})$ over $\Psi^{l@o_{\leftarrow ij}}$, we have
\begin{eqnarray} f_{avg}(\pi^{local}_{ij})-f_{avg}(\pi^{local}_{\leftarrow ij}) &=& \mathbb{E}_{\Psi^{l@o_{\leftarrow ij}}}[\Delta(e^{local} _{ij}\mid \Psi^{l@o_{\leftarrow ij}}_{l_{\leftarrow ij }})]
 \nonumber\\
&\geq& \mathbb{E}_{\Psi^{l@o_{\leftarrow ij}}}[f_{avg}(\pi^{local} @\pi^{opt}_{ij}\mid \Psi^{l@o_{\leftarrow ij}})-f_{avg}(\pi^{local} @\pi^{opt}_{\leftarrow ij}\mid \Psi^{l@o_{\leftarrow ij}})]\label{eq:f}\\
&=& f_{avg}(\pi^{local} @\pi^{opt}_{ij})-f_{avg}(\pi^{local} @\pi^{opt}_{\leftarrow ij}) \label{eq:1}
\end{eqnarray}
where (\ref{eq:f}) is due to (\ref{eq:ddd1}). Because for any policy $\pi$, we have $f_{avg}(\pi)=\sum_{i\in[b], j\in[d_i]}(f_{avg}(\pi_{ij})-f_{avg}(\pi_{\leftarrow ij}))$, it follows that
\begin{eqnarray*}
f_{avg}(\pi^{local})&=&\sum_{i\in[b], j\in[d_i]}(f_{avg}(\pi^{local}_{ij})-f_{avg}(\pi^{local}_{\leftarrow ij}))\\
&\geq& \sum_{i\in[b], j\in[d_i]} (f_{avg}(\pi^{local} @\pi^{opt}_{ij})-f_{avg}(\pi^{local} @\pi^{opt}_{\leftarrow ij}))\\
&=& f_{avg}(\pi^{local} @\pi^{opt}) - f_{avg}(\pi^{local})\\
&\geq&  f_{avg}(\pi^{opt}) - f_{avg}(\pi^{local})
\end{eqnarray*}
The first inequality is due to (\ref{eq:1}), and the second inequality is due to the assumption that $f$ is adaptive monotone.
Then we have $ f_{avg}(\pi^{local}) \geq  f_{avg}(\pi^{opt})/2$. $\Box$

\emph{Remark:} We believe that  Lemma \ref{lem:6}  is of independent interest in the field of adaptive submodular maximization under matroid constraints. In particular,  \citet{golovin2011adaptive1} shows that a classic adaptive greedy algorithm, which iteratively selects an item with the largest contribution to the previously selected items, achieves an approximation ratio $1/2$. This approximation factor is nearly optimal. When applied to our problem, their algorithm needs $O((\sum_{i\in[b]} d_i)n)$ value oracle queries. Since our locally greedy policy $\pi^{local}$  does not rely on any specific ordering of $B_i$, it requires $O(\sum_{i\in[b]} d_i |B_i|)$ value oracle queries.

\subsection{Linear-Time Policy for Fully Adaptive Submodular Functions}
Although $\pi^{local}$ requires less value oracle queries than the classic greedy algorithm, its runtime is still dependent on $d_i$, the cardinality constraint for each group $B_i$. We next present a linear-time adaptive policy \emph{Generalized Adaptive Stochastic Greedy}, denoted $\pi^{gasg}$, whose runtime is independent of $d_i$. However, our analysis requires stronger assumptions about the objective function.

We first introduce the concept of \emph{fully adaptive submodularity}, a stricter condition than the adaptive submodularity.

\begin{definition}[Fully Adaptive Submodularity]
\label{def:1aaa}
For any subset of items $V\subseteq E$ and any integer $a\in[|V|]$, let  $\Omega(V,a)$  denote the set of policies which are allowed to select at most $a$ items only from $V$. A function $f$ is fully adaptive submodular with respect to a prior distribution $ p(\phi)$, if for any two partial realizations $\psi$ and $\psi'$ such that $\psi\subseteq \psi'$, and any subset of items $V\subseteq E$ and any $a\in [|V|]$, the following holds:
\[\max_{\pi\in \Omega(V,a)} \Delta(\pi \mid \psi) \geq \max_{\pi\in \Omega(V,a)} \Delta(\pi \mid \psi') \]
\end{definition}

In the above definition, $V$  can be any single item, thus any fully adaptive submodular function must be adaptive submodular according to Definition \ref{def:11}.

The basic idea of $\pi^{gasg}$ is similar to $\pi^{local}$, except that we leverage the adaptive stochastic greedy policy $\pi^{asg}$ to select a group of $d_i$ items from $B_i$ for each $i\in[b]$. A detailed description of $\pi^{gasg}$ can be found in Algorithm \ref{alg:LPPX}.  We next brief the idea of $\pi^{gasg}$:   Starting with an empty set, $\pi^{gasg}$ first selects $d_1$ number of items from $B_1$ using $\pi^{asg}$, i.e., first sampling a random set $S_{11}$ with size $\frac{|B_1|}{d_1}\log\frac{1}{\epsilon}$ uniformly at random from $B_1$, then selects an item  with the largest conditional expected marginal utility from $S_{11}$, selecting the next item from a newly sampled set  $S_{12}$ in the same greedy manner, and so on; $\pi^{local}$ then selects $d_2$ number of items from $B_2$ using $\pi^{asg}$, where the size of the random set is set to $\frac{|B_2|}{d_2}\log\frac{1}{\epsilon}$,  conditioned on the current observation, and so on. Similar to   $\pi^{local}$,  $\pi^{gasg}$ does not rely on any specific ordering of $B_i$.

\begin{algorithm}[hptb]
\caption{Generalized Adaptive Stochastic Greedy $\pi^{gasg}$}
\label{alg:LPPX}
\begin{algorithmic}[1]
\STATE $A=\emptyset; i=1; j=1$.
\WHILE {$i \leq b$}
\WHILE {$j \leq d_i$}
\STATE observe the current partial realization $\psi^{gasg}_{\leftarrow ij}$;
\STATE $S_{ij}\leftarrow$ a random set, with size $\frac{|B_i|}{d_i}\log\frac{1}{\epsilon}$, sampled  uniformly at random from $B_i$;
\STATE $e^{gasg}_{ij}\leftarrow \arg\max_{e \in S_{ij}}\Delta(e\mid \psi^{gasg}_{\leftarrow ij})$;
\STATE $A\leftarrow A\cup \{e^{gasg}_{ij}\}$; $j\leftarrow j+1$;
\ENDWHILE
$j\leftarrow 1$; $i\leftarrow i+1$;
\ENDWHILE
\RETURN $A$
\end{algorithmic}
\end{algorithm}

We next analyze the performance bound of $\pi^{gasg}$. We start by showing that the expected utility of $\pi^{gasg}$ is at least $\frac{1-1/e-\epsilon}{2-1/e-\epsilon}$ times the expected utility of $\pi^{local}$.
\begin{lemma}
\label{lem:5}
If $f$ is full adaptive submodular and adaptive monotone, then  $f_{avg}(\pi^{gasg}) \geq \frac{1-1/e-\epsilon}{2-1/e-\epsilon} f_{avg}(\pi^{local})$.
\end{lemma}
\emph{Proof:} For ease of presentation, we use $\pi^{local}_i$ (resp. $\pi^{gasg}_i$) to denote a policy that runs $\pi^{local}$ (resp. $\pi^{gasg}$) until it selects all $\sum_{z\in[i]}d_z$ items from $B_1, B_2, \ldots, B_i$. Consider a policy $\pi^{gasg} @\pi^{local}_i$ that runs $\psi^{gasg}$ first, then runs  $\pi^{local}_i$. Let $\psi^{g@l_i}$ denote the partial realization obtained after running $\pi^{gasg} @\pi^{local}_i$. We use $\psi^{g@l_i}_{g_i}\subseteq \psi^{g@l_i}$ to denote a subrealization of $\psi^{g@l_i}$ obtained after running $\pi^{gasg}_i$.   For a fixed $i\in[b]$, let $\Omega(B_i, d_i)$ denote a set of polices which are allowed to select at most $d_i$ items only from $B_i$. For a given $\psi^{g@l_i}$, we first bound the conditional expected marginal utility $\Delta(\pi^{gasg}_{i+1}\mid \psi^{g@l_i}_{g_i})$ of the $d_{i+1}$ items  selected  from $B_{i+1}$ by $\pi^{gasg}$. Let $\alpha=1-1/e-\epsilon$,
\begin{eqnarray}
\Delta(\pi^{gasg}_{i+1}\mid \psi^{g@l_i}_{g_i})&\geq&\alpha\max_{\pi\in \Omega(B_{i+1}, d_{i+1})}\Delta(\pi \mid \psi^{g@l_i}_{g_i})\nonumber\\
&\geq& \alpha\max_{\pi\in \Omega(B_{i+1}, d_{i+1})} \Delta(\pi \mid \psi^{g@l_i})\nonumber\\
&\geq& \alpha \Delta(\pi^{gasg} @ \pi^{local}_{i+1} \mid \psi^{g@l_i})\nonumber\\
&=& \alpha (f_{avg}(\pi^{gasg} @\pi^{local}_{i+1}\mid  \psi^{g@l_i})-f_{avg}(\pi^{gasg} @\pi^{local}_{i}\mid  \psi^{g@l_i})) \label{eq:2}
\end{eqnarray}
The first inequality is due to Theorem \ref{thm:main} and the fact that we use $\pi^{asg}$ to select $d_i$ items from each group $B_i$. The second inequality is due to $f$ is adaptive monotone.

 Let $\Psi^{g@l_i}$ denote a random partial realization obtained after running $\pi^{gasg} @\pi^{local}_i$. Taking the expectation of $\Delta(\pi^{gasg}_{i+1}\mid \Psi^{g@l_i})$ over $\Psi^{g@l_i}$,  we have
\begin{eqnarray}
f_{avg}(\pi^{gasg}_{i+1})-f_{avg}(\pi^{gasg}_{i}) &=&  \mathbb{E}_{\Psi^{g@l_i}}[\Delta(\pi^{gasg}_{i+1}\mid \Psi^{g@l_i}_{g_i})]\nonumber\\
&\geq& \mathbb{E}_{\Psi^{g@l_i}}[ \alpha (f_{avg}(\pi^{gasg} @\pi^{local}_{i+1}\mid  \Psi^{g@l_i})-f_{avg}(\pi^{gasg} @\pi^{local}_{i}\mid  \Psi^{g@l_i}))]\nonumber\\
&=& \alpha (f_{avg}(\pi^{gasg} @\pi^{local}_{i+1})-f_{avg}(\pi^{gasg} @\pi^{local}_{i}))\label{eq:3}
\end{eqnarray}
The inequality is due to (\ref{eq:2}).

Because $f_{avg}(\pi^{gasg})=\sum_{i\in[b-1]} (f_{avg}(\pi^{gasg}_{i+1})-f_{avg}(\pi^{gasg}_{i}))$ and $f_{avg}(\pi^{gasg} @\pi^{local})=f_{avg}(\pi^{gasg}) + \sum_{i\in[b-1]}(f_{avg}(\pi^{gasg} @\pi^{local}_{i+1})-f_{avg}(\pi^{gasg} @\pi^{local}_{i}))$, we have
\begin{eqnarray*}
f_{avg}(\pi^{gasg})&=&\sum_{i\in[b-1]} (f_{avg}(\pi^{gasg}_{i+1})-f_{avg}(\pi^{gasg}_{i})) \\
&\geq& \alpha\sum_{i\in[b-1]} (f_{avg}(\pi^{gasg} @\pi^{local}_{i+1})-f_{avg}(\pi^{gasg} @\pi^{local}_{i}))\\
&=&\alpha ( f_{avg}(\pi^{gasg} @\pi^{local})-f_{avg}(\pi^{gasg}))\\
&\geq& \alpha ( f_{avg}(\pi^{local})-f_{avg}(\pi^{gasg}))
\end{eqnarray*}
The first inequality is due to (\ref{eq:3}) and the second inequality is due to $f$ is adaptive monotone. It follows that $f_{avg}(\pi^{gasg}) \geq \frac{\alpha}{1+\alpha}f_{avg}(\pi^{local})=\frac{1-1/e-\epsilon}{2-1/e-\epsilon}f_{avg}(\pi^{local})$. $\Box$

We next present the main theorem and show that the approximation ratio of $\pi^{gasg}$ is at least $ \frac{1-1/e-\epsilon}{4-2/e-2\epsilon}$ and its running time is $O(n\log\frac{1}{\epsilon})$ (number of value oracle queries).
\begin{theorem}
If $f$ is fully adaptive submodular and adaptive monotone, then  $f_{avg}(\pi^{gasg}) \geq \frac{1-1/e-\epsilon}{4-2/e-2\epsilon} f_{avg}(\pi^{opt})$, and $\pi^{gasg}$ uses at most $O(n\log\frac{1}{\epsilon})$ value oracle queries.
\end{theorem}

\emph{Proof:} Lemma \ref{lem:6} and Lemma \ref{lem:5} together imply that $f_{avg}(\pi^{gasg}) \geq \frac{1-1/e-\epsilon}{4-2/e-2\epsilon} f_{avg}(\pi^{opt})$. Recall that in Algorithm \ref{alg:LPPX}, we set the size of the random set  $S_{ij}$ to $\frac{|B_i|}{d_i}\log\frac{1}{\epsilon}$, thus the total number of value oracle queries is $\sum_{i\in[b], j\in{[d_i]}}\frac{|B_i|}{d_i}\log\frac{1}{\epsilon}= n\log\frac{1}{\epsilon}$. $\Box$
\section{Conclusion}
In this paper, we study the non-monotone adaptive submodular maximization problem subject to a cardinality constraint. We first revisit the adaptive random algorithm and show that it achieves a $1/e$ approximation ratio under adaptive submodularity. Then we propose a linear-time adaptive policy that achieves a $1/e-\epsilon$ approximation ratio, using only $O(n\epsilon^{-2}\log \epsilon^{-1})$ value oracle queries. Then, we propose a faster algorithm for the monotone case, where our algorithm  achieves a $1-1/e-\epsilon$ approximation ratio in expectation  with $O(n \log  \frac{1}{\epsilon})$ value oracle queries. For the monotone case, we generalize our study by considering a single partition matroid constraint, and develop a linear-time algorithm for fully adaptive submodular functions. In the future, we would like to consider other constraints such as knapsack constraints and general matroid constraints.

\section{Appendix}
\subsection{Proof of Lemma \ref{lem:2222}}
 Let $v_1(\psi^{{r-1}}), v_2(\psi^{{r-1}}), \ldots , v_k(\psi^{{r-1}})$ be the $k$ items with the largest marginal contribution to $\psi^{{r-1}}$, sorted in a non-increasing order of their marginal contributions. We use  $X_j$ to denote an indicator for the event $e'_r = v_j(\psi^{{r-1}})$. The following two lemmas are proved in \citep{buchbinder2017comparing}.
\begin{lemma}
\label{lem:9999}
For any $j \in[k]$, let $X_j=1$ if $v_j(\psi^{{r-1}})$ is picked as $e'_r$, and  $X_j=0$ otherwise, we have $\mathbb{E}[\sum_{j\in[k]}X_j]\geq 1-\epsilon$.
\end{lemma}

\begin{lemma}
\label{lem:8888}
The probability $\mathbb{E}[X_j]$ that  $v_j(\psi^{{r-1}})$ is picked as $e'_r$ is a non-increasing function of $j$.
\end{lemma}
Now we are ready to prove the lemma. Our proof generalizes the proof of Lemma 4.5 in \citep{buchbinder2017comparing} to the adaptive setting. Let  $\Psi^{{r-1}}$ denote a random partial realization observed after running $\pi^{lt}_{r-1}$. The expectation $\mathbb{E}_{ \Psi^{{r-1}}}[\cdot]$ is taken over all such partial realizations $\Psi^{{r-1}}$. Then we have
\begin{eqnarray}
f_{avg}(\pi^{lt}_r) - f_{avg}(\pi^{lt}_{r-1}) &=& \mathbb{E}_{ \Psi^{{r-1}}}[\mathbb{E}_{e'_r}[\max\{\Delta(e'_r \mid\Psi^{{r-1}}),0\}]]\\
&\geq&\mathbb{E}_{ \Psi^{{r-1}}}[\sum_{j\in[k]}\mathbb{E}[X_j]\max\{\Delta(v_j(\Psi^{{r-1}}) \mid\Psi^{{r-1}}),0\}]\\
&\geq& \mathbb{E}_{ \Psi^{{r-1}}}[\frac{\sum_{j\in[k]}\mathbb{E}[X_j] \sum_{j\in[k]} \max\{\Delta(v_j(\Psi^{{r-1}}) \mid\Psi^{{r-1}}),0\}}{k}]\\
&\geq& \mathbb{E}_{ \Psi^{{r-1}}}[\frac{\sum_{j\in[k]}\mathbb{E}[X_j] \Delta(\pi^{opt}\mid \Psi^{{r-1}})}{k}]\\
&\geq& (1-\epsilon)\mathbb{E}_{ \Psi^{{r-1}}}[\frac{ \Delta(\pi^{opt}\mid \Psi^{{r-1}})}{k}]\\
&=& (1-\epsilon)\frac{f_{avg}(\pi^{opt}@\pi^{lt}_{r-1})-f_{avg}(\pi^{lt}_{r-1})}{k}
\end{eqnarray}
The second inequality is due to Chebyshev's sum inequality because $\max\{\Delta(v_j(\Psi^{{r-1}}) \mid\Psi^{{r-1}}),0\}$ is nonincreasing
in $j$ by definition and $\mathbb{E}[X_j]$ is a non-increasing function of $j$ by Lemma \ref{lem:8888}, the third inequality is due to $f$ is adaptive submodular and Lemma 1 in \citep{gotovos2015non}, the forth inequality is due to Lemma \ref{lem:9999}.

\subsection{Proof of Lemma \ref{lem:aX}}   We first provide a lower bound on the probability that $R\cap  A^* =\emptyset$.
\begin{eqnarray}
\Pr[R\cap  A^* = \emptyset]&=&{(1-\frac{|A^*|}{|E|})}^{|R|} ={(1-\frac{k}{n})}^{|R|}\leq e^{-|R|\frac{k}{n}}
\end{eqnarray}

It follows that $\Pr[R\cap  A^* \neq \emptyset]\geq 1- e^{-\frac{|R|}{n}k}$. Because we assume $|R|=\frac{n}{k}\log\frac{1}{\epsilon}$, we have \begin{equation}\label{eq:a}\Pr[R\cap  A^*\neq \emptyset]\geq 1-e^{-|R|\frac{k}{n}} \geq 1-  e^{-\frac{\frac{n}{k}\log\frac{1}{\epsilon}}{n}k}\geq 1-\epsilon
\end{equation}

\bibliographystyle{ijocv081}
\bibliography{reference}




\end{document}